\documentclass[journal=eds]{CUP-JNL-DTM}%

\usepackage{graphicx}
\usepackage{multicol,multirow}
\usepackage{amsmath,amssymb,amsfonts}
\usepackage{mathrsfs}
\usepackage{amsthm}
\usepackage{rotating}
\usepackage{appendix}
\usepackage{ifpdf}
\usepackage[T1]{fontenc}
\usepackage{newtxtext}
\usepackage{newtxmath}
\usepackage{textcomp}
\usepackage{xcolor}
\usepackage{lipsum}
\usepackage[colorlinks,allcolors=blue]{hyperref}
\usepackage{cleveref}

\theoremstyle{definition}

\numberwithin{equation}{section}

\jname{Environmental Data Science}
\articletype{APPLICATION PAPER}
\jyear{YEAR}

\begin{document}

\begin{Frontmatter}

\title[When Bigger is Worse]{When Bigger is Worse: A Practitioner's Guide to Model Selection Under Data Scarcity}

\author[1]{Kwame Mbobda-Kuate}
\author[2]{Gabriel Kasmi}

\authormark{Kwame Mbobda-Kuate and Gabriel Kasmi}

\address[1]{
  \orgname{ENSAE Paris},
  \orgaddress{\city{Palaiseau}, \postcode{91120}, \country{France}}.
  \email{firstname.lastname@ensae.fr}}

\address[2]{
  \orgdiv{Centre O.I.E.},
  \orgname{Mines Paris -- PSL University},
  \orgaddress{\city{Sophia-Antipolis}, \postcode{06904}, \country{France}}.
  \email{firstname.lastname@minesparis.psl.eu}}

\authormark{Kwame Mbobda-Kuate and Gabriel Kasmi}




\abstract{Scaling laws assume larger models trained on more data consistently 
outperform smaller ones — an assumption that drives model selection 
in computer vision but remains untested in resource-constrained Earth 
observation (EO). We conduct a systematic efficiency analysis across 
three scaling dimensions: model size, dataset size, and input resolution, 
on rooftop photovoltaic (PV) detection in Madagascar, yielding 180 training runs across 
60 configurations. Optimizing for \textit{model efficiency} 
(mAP$_{50}$ per unit of model size), we find a consistent efficiency 
inversion: YOLO11N achieves the highest efficiency ($22\times$ higher than YOLO11X) 
with no accuracy penalty: it reaches the second highest absolute 
mAP$_{50}$ (0.459), outperforming all models except YOLO11S by a margin 
smaller than run-to-run variance, 
directly contradicting the scaling prior. Resolution is the dominant 
resource allocation lever: moving from 416\,px to 1280\,px at 10\% of 
the data matches the efficiency gain of collecting the full dataset at 
low resolution. These findings are robust to the deployment objective: 
small high-resolution configurations are Pareto-dominant across all 60 
experimental setups in the joint accuracy--throughput space, leaving no 
tradeoff to resolve. In data-scarce EO, the scaling prior does not just 
fail: it inverts.}

\end{Frontmatter}

\section*{Impact Statement}
Earth observation practitioners default to the largest model their compute budget allows, following a scaling prior imported from data-rich domains like language processing. This paper shows that in data-scarce Earth observation settings --- the norm, not the exception, in developing countries --- this default hurts both accuracy and deployment cost. On rooftop PV detection in Madagascar, the smallest YOLO11 variant matches or beats every larger model while using 22× less storage. There is no tradeoff to resolve. The result also redirects a scarce resource: annotation budget. Since resolution beats data volume, teams operating under tight annotation constraints should invest in higher-resolution acquisition before more labeling. The mechanism (overparameterization ratio, absence of COCO domain transfer) likely generalizes to other small-object EO tasks under annotation scarcity.


\section{Introduction}\label{sec:intro}
Deep learning progress in computer vision has been largely driven by scaling: larger models 
trained on larger datasets consistently push benchmark performance upward. This paradigm, 
formalized through scaling laws~\citep{kaplan2020scaling,hoffmann2022training}, has proven 
remarkably predictable in data-rich settings such as natural language processing (NLP) and large-scale image 
classification~\citep{dosovitskiy2021image,zhai2022scaling}. Yet this predictability rests on the underlying assumption of data abundance.

Earth observation (EO) challenges this assumption on two fronts. First, EO datasets are 
structurally data-scarce: annotation is expensive, requires domain expertise, and is 
geographically uneven~\citep{cheng2020remote}. Second, deployment in operational settings, 
e.g., in developing countries where most renewable energy growth will 
occur~\citep{iea2023world}, imposes hard compute constraints that make large models 
impractical regardless of their accuracy. In this regime, the relevant criterion is not raw 
performance but \textit{efficiency}, i.e., the performance per unit of resource. Moreover, since pretrained representations provide no exploitable prior for domain-specific EO 
tasks~\citep{yosinski2014transferable}, larger models cannot rely on richer pretraining to 
compensate for data scarcity.

Despite growing evidence that architectural innovation can match or exceed naive parameter 
scaling~\citep{tan2019efficientnet,frankle2018lottery,shi2024when,beyer2022better}, and that 
smaller models outperform larger ones below certain data 
thresholds~\citep{brigato2021close}, no systematic analysis has jointly characterized the 
interplay between model size, dataset fraction, and input resolution in a data-scarce EO 
setting. Existing work addresses these dimensions in isolation --- resolution 
effects~\citep{shermeyer2019effects}, data-scarce regimes~\citep{brigato2021close}, or 
budget-aware detection~\citep{pardo2021baod} --- but their joint optimization under 
operational constraints remains an open question.

This paper addresses that gap through a systematic efficiency analysis of the YOLO11 
\citep{yolo11_ultralytics} model family on rooftop photovoltaic (PV) detection in Madagascar --- a representative data-scarce, small object EO task in a developing country context. We 
evaluate five model variants across four dataset fractions and three input resolutions. Following~\cite{tan2019efficientnet}, we define efficiency as the mean averave precision (mAP$_{50}$) per unit of model size (in MB). As illustrated in 
Figure~\ref{fig:pareto-efficiency}, our analysis reveals a striking result: the smallest 
model (YOLO11N, 2.6M parameters) achieves the second highest absolute mAP$_{50}$ (0.459, 
indistinguishable from the best-performing YOLO11S within run-to-run variance) and a 
$22\times$ efficiency advantage over the largest variant. Moreover, this model lies at the apex of the 
accuracy--throughput Pareto frontier, directly contradicting the scaling prior. We further 
show that resolution is the dominant resource allocation lever, and that this recommendation 
is robust to the choice of deployment objective: small high-resolution configurations 
simultaneously maximize detection accuracy and inference throughput, leaving no 
accuracy--speed tradeoff to resolve. We provide a mechanistic explanation grounded in the 
bias-variance tradeoff, compute-optimal training theory~\citep{vapnik2013nature,hoffmann2022training}, 
and the structural absence of domain transfer from large-scale pretraining in this setting. The key results of this work are:
\begin{itemize}
    \item A systematic efficiency analysis across three scaling dimensions (model size, dataset 
    fraction, input resolution) in a data-scarce EO setting, yielding 180 training runs across 
    60 configurations on a real-world rooftop PV dataset.
    \item Evidence that smaller models dominate larger ones in both accuracy and efficiency 
    under data scarcity, with a $22\times$ efficiency gap between YOLO11N and YOLO11X and 
    no statistically meaningful accuracy penalty.
    \item A mechanistic explanation of this inversion grounded in overparameterization theory, 
    supported by empirical learning curves and overparameterization ratio analysis.
    \item A practical resource allocation recommendation: prioritize input resolution over 
    dataset volume, and select the smallest model consistent with task requirements.
    \item A joint efficiency--throughput analysis showing that small high-resolution 
    configurations are Pareto-dominant across, confirming 
    that the resource allocation recommendation holds regardless of whether the practitioner 
    prioritizes static efficiency or real-time deployment.
\end{itemize}

Code for replicating the results of this paper can be accessed at the following URL : \url{https://github.com/kwame-mbobda-kuate/scaling-laws-eds}.
\section{Related works}\label{sec:related}
\subsection{Scaling Laws and Model Efficiency}

A classical assumption underpinning the scaling paradigm is that larger 
models consistently outperform smaller ones across domains~(\cite{kaplan2020scaling}). 
This assumption has driven state-of-the-art progress in NLP, where models 
are trained in a self-supervised manner on colossal amounts of text --- a 
setting fundamentally different from the data-scarce, domain-specific regime 
we consider. Several lines of work challenge its generality in vision. The 
lottery ticket hypothesis~(\cite{frankle2018lottery}) shows that sparse subnetworks 
within large models can match full model performance, suggesting substantial 
parameter redundancy. \cite{tan2019efficientnet} demonstrate that compound 
scaling outperforms naive parameter scaling and introduced \textit{model 
efficiency}, i.e., detection performance per unit of model size, as the 
operationally relevant criterion for resource-constrained deployment, a 
definition we adopt directly. Vision transformer scaling studies further 
reveal diminishing returns at larger scales~(\cite{beyer2022better}), 
suggesting that the relationship between model size and performance is 
neither monotonic nor universal. More broadly, \cite{zhai2022scaling} show 
that these diminishing returns are not confined to specific architectures 
but reflect a structural property of the scaling regime itself --- one that 
breaks down precisely when the data abundance assumption fails.

Two distinct notions of efficiency recur in this literature and are worth 
distinguishing explicitly. \textit{Model efficiency} (mAP per unit of model 
size or parameters) captures the static deployment cost --- the criterion 
relevant when memory footprint constrains model selection on edge devices. 
\textit{Inference throughput} (mAP per unit of inference time, typically 
visualized as accuracy--FPS Pareto frontiers) captures the dynamic deployment 
cost --- the criterion relevant when latency constrains real-time operation. 
These two notions are correlated but not equivalent: a model with a small 
memory footprint may not be the fastest at inference, particularly when input 
resolution increases GFLOPs independently of parameter count. Most prior work 
optimizes for one criterion implicitly without acknowledging the other; we 
address both explicitly and show that, in our setting, the same configurations 
dominate on both axes simultaneously.

\cite{shi2024when} show that small vision models augmented with multi-scale 
features can surpass much larger counterparts on classification and detection 
tasks in terms of both model efficiency and raw accuracy, questioning the need 
for large models beyond a certain data regime. Their Scaling on Scales method 
demonstrates that injecting multi-scale spatial context --- effectively 
increasing the resolution of the signal available to the model --- recovers 
the performance gap between small and large models at a fraction of the 
parameter cost. This is consistent with our finding that input resolution, 
rather than model capacity, is the dominant resource allocation lever: in 
both cases, richer spatial information substitutes for additional parameters.

\subsection{Data-Scarce Regimes and Resource-Constrained Learning}

Traditional scaling laws assume data abundance, an assumption that breaks 
down in specialized domains with limited annotation budgets. Few-shot 
learning addresses this by exploiting architectural priors to generalize 
from few examples~(\cite{snell2017prototypical}), while meta-learning extends 
this to rapid domain adaptation~(\cite{finn2017maml}). Transfer learning offers 
a complementary strategy, though its effectiveness varies significantly with 
the domain gap between pretraining and target tasks~(\cite{yosinski2014transferable}), 
a limitation directly relevant to our setting, where COCO-pretrained models 
face an out-of-distribution target domain (see \cref{sec:zero_shot}).

Beyond architectural solutions, recent work has examined the optimal 
allocation of a fixed resource budget. \cite{brigato2021close} demonstrate 
empirically that smaller models outperform larger ones below a dataset-size 
threshold --- 160 samples per class on CIFAR-10 and FashionMNIST, 80 on 
Street View House Numbers --- establishing a crossover regime that directly 
motivates our analysis. This result, however, is established on image 
classification benchmarks; its transposition to object detection in EO 
involves an additional gap, as detection tasks introduce localization 
objectives and class imbalance that may shift the crossover threshold. 
\cite{hoffmann2022training} formalize compute-optimal training, showing 
that model size and data volume must scale proportionally for optimal 
performance (with a token~/~parameters ratio of 20:1) --- a model 
undertrained relative to its capacity is systematically suboptimal. 
Approximating tokens as non-overlapping $16\times16$ image patches, this 
rule predicts that YOLO11N (2.6M parameters) is near the compute-optimal 
regime for our training set, while YOLO11X (57M parameters) operates at 
a~20$\times$ deficit --- a prediction our efficiency results confirm 
empirically, and one that extends the Chinchilla insight from LLM 
pretraining to supervised fine-tuning under annotation constraints. 
In the object detection setting specifically, \cite{pardo2021baod} 
introduce Budget-Aware Object Detection (BAOD), showing that jointly 
optimizing annotation strategy under a fixed budget matches 
fully-supervised performance while reducing annotation cost by 12.8\%. 
Where BAOD treats model size as fixed and optimizes the annotation 
strategy, we hold the annotation scheme fixed and jointly optimize 
model size, data volume, and input resolution --- extending the 
budget-aware framing to the model selection dimension.

\subsection{Small Object Detection in Remote Sensing}

Small object detection presents challenges that are orthogonal to model 
scaling: limited feature representation at low resolutions, high intra-class 
scale variation, and domain shift between classification pretraining and 
detection fine-tuning~(\cite{singh2018analysis}). Aerial and satellite imagery 
compounds these difficulties through viewpoint variation, occlusion, and 
background clutter~(\cite{cheng2016survey}). Architectural responses include 
feature pyramid networks~(\cite{lin2017feature}) and multi-scale sampling 
strategies~(\cite{singh2018sniper}), while context-aware augmentation has been 
shown to compensate for training sample scarcity~(\cite{dvornik2018modeling}).

Critically for our setting, resolution scaling has been shown to be more 
effective than model scaling for small object detection in satellite imagery. 
\cite{shermeyer2019effects} demonstrate that increasing input resolution yields 
substantial gains in aerial object detection performance, even when applied 
through super-resolution techniques. They find that performance degrades from 
mAP~=~0.53 at 30~cm/pixel resolution to mAP~=~0.11 at 4.8~m/pixel resolution, while 
super-resolving native 30~cm/pixel imagery to 15~cm/pixel yields the greatest benefit: a 
13--36\% improvement in mAP. These findings directly motivate our joint 
analysis of resolution and model size: if resolution alone drives such gains 
independently of architectural choices, then resolution should be the primary 
resource allocation lever --- a hypothesis our experiments confirm and 
quantify under controlled data-scarce conditions.

\subsection{Rooftop PV Detection from Satellite Imagery}

Rooftop PV detection from satellite imagery is well-established in 
developed countries, where high-resolution imagery is readily available. 
Existing works leverage CNN and ViT-based architectures across 
France~(\cite{Kasmi2022_TowardsUnsupervisedAssessment}), 
Germany~(\cite{Mayer2022_3DPVLocator}), the 
Netherlands~(\cite{Kausika2021_GeoAI_Netherlands}), and the United 
States~(\cite{yu2018solar}). These works primarily optimize for accuracy 
under data abundance, with only limited attention to inference and 
deployment cost~(\cite{parhar2022hyperionsolarnet,kasmi_enhancing_2024}).

Developing countries have received comparatively little attention, despite 
cumulating high solar potential and operating less resilient electric grids 
where unmonitored PV installations carry greater grid stability 
risks~(\cite{iea2023world}). PV mapping in these contexts remains largely 
unexplored, with only isolated case studies in 
Tunisia~(\cite{bouaziz2024high}). This gap motivates our choice of the 
OpenStat Madagascar dataset~(\cite{OpenStat}): it provides a realistic 
data-scarce benchmark with limited annotation budgets, heterogeneous 
imagery, and deployment under compute constraints --- a natural testbed 
for our efficiency analysis.

The works reviewed above address complementary dimensions of the 
efficiency problem, but always in isolation: scaling 
laws~(\cite{kaplan2020scaling,hoffmann2022training}) operate in 
data-abundant NLP settings; data-scarce regime 
studies~(\cite{brigato2021close}) focus on image classification rather 
than detection; budget-aware detection~(\cite{pardo2021baod}) treats model 
size as fixed; and resolution studies~(\cite{shermeyer2019effects}) vary 
input resolution independently of model complexity and data volume. No 
existing work jointly characterizes the interplay between model size, 
dataset fraction, and input resolution in a data-scarce Earth observation 
setting, leaving practitioners with no principled guidance on how to 
allocate a fixed resource budget across these three dimensions 
simultaneously. This paper addresses that gap.
\section{Methods}\label{sec:methods}
Our aim is to systematically investigate the effects of model size, input resolution, and 
training dataset size on detection performance under resource constraints. Specifically, we 
address the following research question: \textit{given a fixed resource budget, what is the 
optimal allocation between model complexity, data volume, and image resolution for small object 
detection in a data-scarce Earth observation setting?} Throughout our experiments, we report 
computational cost alongside accuracy metrics, and define our measure of interest as efficiency 
rather than pure performance.

We focus on efficiency rather than pure accuracy because, in resource-constrained deployment 
contexts, a model that achieves marginally higher detection performance at disproportionate 
computational cost is not operationally viable. Efficiency captures performance per unit of 
resource --- the relevant criterion for practitioners operating under hardware constraints, 
a setting representative of Earth observation applications in developing countries.

\subsection{Experimental Setup}

We conduct systematic scaling experiments across three dimensions: model complexity, dataset 
fraction, and input resolution, producing a factorial grid of training configurations.

\paragraph{Model Complexity.}
We evaluate five YOLO11~(\cite{yolo11_ultralytics}) variants with increasing capacity, spanning 
a $22\times$ range in parameter count and a $21\times$ range in model size (Table~\ref{tab:models}). 
We select YOLO11 for three reasons. First, it is among the most widely deployed single-stage 
detector families in operational EO pipelines, making our results directly actionable for 
practitioners. Second, all five variants share a common architecture and differ only in 
capacity, allowing us to isolate the effect of model size without confounding architectural 
choices. Third, the $22\times$ size range provides sufficient resolution to characterize 
scaling behavior across the full spectrum from edge-deployable to server-class models.

\begin{table}[h]
\centering
\caption{\textbf{YOLO11 model variants evaluated in this study}}
\begin{tabular}{lcc}
\toprule
\textbf{Variant} & \textbf{Parameters} & \textbf{Size (MB)} \\
\midrule
YOLO11N & 2.6M  & 5.1   \\
YOLO11S & 9.5M  & 18.1  \\
YOLO11M & 20.1M & 38.4  \\
YOLO11L & 25.4M & 48.5  \\
YOLO11X & 57.0M & 108.6 \\
\bottomrule
\end{tabular}
\label{tab:models}
\end{table}

\paragraph{Dataset Fraction.}
Each model is trained on four fractions of the available training set: 10\% (897 images), 
25\% (2{,}244 images), 50\% (4{,}488 images), and 100\% (8{,}977 images). This design allows 
us to characterize model behavior across a realistic range of annotation budgets, from 
severely data-scarce to fully supervised.

\paragraph{Input Resolution.}
We evaluate three input resolutions: 416\,px, 640\,px, and 1280\,px. These correspond to 
the standard production configurations recommended by Ultralytics, ensuring that our results 
are directly comparable to practitioner deployments. Resolution directly affects the 
visibility of small rooftop PV installations and constitutes an independent axis of the 
resource budget: higher resolution increases both GPU memory consumption and inference time, 
independently of model size.

\paragraph{Training Grid.}
These settings produce $5 \times 4 \times 3 = 60$ nominal combinations, each reproduced 
with three different random seeds for robustness, yielding 180 training runs in total.

\paragraph{Hyperparameters.}
All runs share identical hyperparameters recommended by Ultralytics to ensure fair comparison: 
50 epochs, default data augmentations, nominal batch size 64 (the actual batch size varies 
but gradient accumulation is used in any case), Adam optimizer (with learning rate of 
$10^{-3}$), and a fixed set of random seeds. All models are initialized from COCO-pretrained 
weights. Hyperparameters and data augmentations are detailed in 
\cref{sec:implementational-details}. While we do not use early stopping, the reported 
metrics correspond to the checkpoint achieving the best validation performance.

\subsection{Evaluation Metrics}

Following~\cite{tan2019efficientnet}, we define model efficiency as the ratio of detection 
performance to model size:

\[
\mathrm{Eff} = 10\times\frac{\mathrm{mAP}_{50}}{\text{ModelSize (MB)}}
\]

Higher values indicate better performance per unit of storage cost. The scaling factor of 10 
is chosen so that the efficiency score is interpretable around unity: a model achieving 
mAP$_{50} = 0.5$ with a 5\,MB footprint yields $\mathrm{Eff} = 1.0$, providing a natural 
break-even reference. We adopt model size (MB) rather than FLOPs as the denominator for 
three reasons: (1) model size is a fixed, resolution-independent quantity, whereas FLOPs 
vary with input resolution and would confound the experimental design; (2) model size 
directly captures the deployment cost most relevant to practitioners selecting models for 
memory-constrained edge devices --- it is the quantity they read off a model card before 
deployment; (3) model size is resolution-invariant, making it the appropriate denominator 
for an experiment that explicitly varies resolution as an independent axis.

We acknowledge that model size is an imperfect proxy for computational cost: a model twice 
as large does not necessarily require twice the inference time. However, our conclusions are 
robust to this choice: \cref{sec:pareto} independently confirms the same ranking using a 
throughput-based criterion (FPS), and \cref{sec:additional-metrics} shows that the efficiency 
ordering is invariant to the choice of numerator metric. Secondary metrics including 
mAP$_{50\text{-}95}$, precision, and recall are reported in \cref{sec:additional-metrics} 
for completeness.
\section{Data}\label{sec:data}
\subsection{OpenStat Madagascar Dataset}

The OpenStat Madagascar dataset~\citep{OpenStat} is a crowdsourced collection of annotated 
imagery of rooftop photovoltaic installations across Madagascar, curated by the Madagascar 
Initiatives for Digital Innovation (MAIDI) with support from the Lacuna Fund. Images were 
acquired between July and September 2023 through two modalities: satellite screenshots 
captured via Google Earth ($\simeq 700\times400$\,px) and drone imagery (ranging from 
$3840\times2160$\,px to $5280\times3956$\,px). The annotation methodology combined visual 
identification on satellite imagery with 25\% ground-truth field verification, followed by 
manual polygon annotation of individual solar objects using GIMP. The full dataset contains 
130,500 annotated solar objects across 11,154 images corresponding to 8,454 distinct 
geolocations. Object types include rooftop solar panels (85.5\%), solar boilers (10.5\%), 
mixed installations (3.8\%), and solar parks ($<$0.1\%).

\begin{figure}[h]
    \centering
    \includegraphics[width=0.5\linewidth]{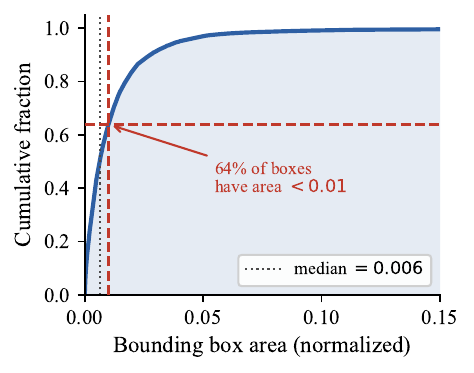}
    \caption{\textbf{Cumulative distribution of bounding box areas (normalized by image area) 
    across the training set.} The median normalized area is 0.6\% of the image area, and 64\% of boxes 
    fall below the 1\% threshold commonly used to define small objects in detection 
    benchmarks~(\cite{lin2017feature}). This confirms that rooftop PV detection in drone 
    imagery is structurally a small object task, justifying the central role of input 
    resolution in our efficiency analysis}
    \label{fig:bbox_cdf}
\end{figure}

This dataset fills a notable gap in the PV detection literature. Existing labeled 
datasets cover France~\citep{kasmi2023crowdsourced} or the United 
States~\citep{bradbury2016distributed,furedi2026labeled}, and worldwide coverage~\citep{kruitwagen2021global,li2025global} 
is limited to PV plants. These datasets rely on satellite or aerial imagery at fixed 
resolution, with annotations collected under controlled conditions and expert supervision. 
OpenStat Madagascar is, to our knowledge, the first large-scale PV detection dataset based 
predominantly on drone imagery (90\% of images), providing substantially higher spatial 
resolution than satellite-based counterparts. As shown in Figure~\ref{fig:bbox_cdf}, 64\% 
of annotated bounding boxes have a normalized area below 0.01, confirming the small object 
nature of the detection task and directly motivating our focus on resolution as a resource 
allocation lever.

\subsection{Dataset Construction and Splits}

For model training, we restrict the dataset to drone images and exclude images containing 
no solar panels, reducing the dataset from 11,154 to 8,977 images. Satellite images are 
excluded due to their substantially lower resolution, which would conflate resolution 
effects with modality effects in our scaling analysis. We convert polygon annotations to 
axis-aligned bounding boxes for compatibility with YOLO11. Among the retained images, 
8,832 (98.4\%) contain only rooftop solar panels and 145 (1.6\%) contain both solar panels 
and boilers.

The dataset is split geographically at the city level: cities are randomly 
assigned to training (80\%, 7,180 images), validation (10\%, 895 images), and test (10\%, 902 images), ensuring 
that no city appears in more than one split. This design enforces spatial 
disjointness between partitions, eliminating the spatial autocorrelation 
that would arise from a random image-level split --- a known source of 
optimistic bias in EO benchmarks~(\cite{rustowicz2019semantic}). Stratified 
sampling ensures that the class balance and geographic coverage are preserved 
across splits. The geographic distribution of images across splits is 
presented in Appendix~\ref{sec:additional_dataset}, \cref{fig:localization}: 
sampling points are concentrated along major transportation corridors and 
around Antananarivo, with secondary clusters along the east coast and in 
the southern regions around Fianarantsoa and Toliara. Annotations were 
produced through crowdsourcing rather than expert labeling, imagery sources 
and acquisition conditions are heterogeneous, and geographic coverage 
reflects access constraints rather than systematic sampling --- making this 
a realistic, if challenging, benchmark representative of the data quality 
faced by practitioners in developing countries.
\section{Results}
\begin{figure}[h]
    \centering
    \includegraphics[width=0.9\linewidth]{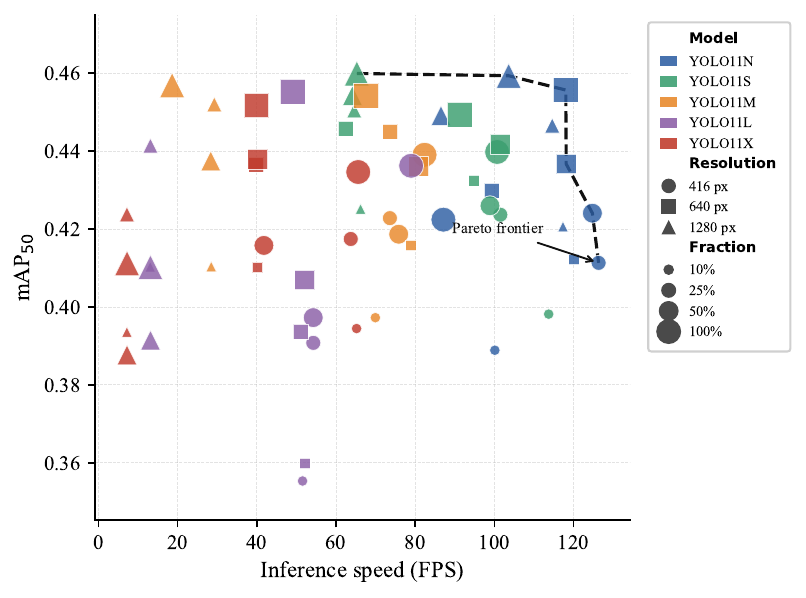}
    \caption{\textbf{Small models at high resolution dominate across all 
    deployment objectives.} Detection performance (mAP$_{50}$) versus 
    inference speed (FPS) for all 180 experimental runs (60 configurations 
    $\times$ 3 seeds) on the OpenStat Madagascar rooftop PV dataset 
    \citep{OpenStat}. Marker shape encodes input resolution (circle: 
    416\,px, square: 640\,px, triangle: 1280\,px) and marker size encodes 
    the training dataset fraction (10\% to 100\%). The dashed line 
    delineates the Pareto frontier. YOLO11N and YOLO11S at 1280\,px lie 
    at the apex, simultaneously maximising accuracy and throughput with 
    no accuracy--speed tradeoff to resolve}
    \label{fig:pareto-efficiency}
\end{figure}

\subsection{No Tradeoff to Resolve: Small Models at High Resolution Dominate}\label{sec:pareto}

Figure~\ref{fig:pareto-efficiency} plots mAP$_{50}$ against inference speed (FPS) 
across all 60 configurations. The result is unambiguous: YOLO11N and YOLO11S at 
1280\,px jointly dominate the Pareto frontier, simultaneously maximizing accuracy 
and throughput. No configuration achieves higher mAP$_{50}$ or higher FPS by 
selecting a larger model or a lower resolution. The scaling prior predicts a 
tradeoff; the data show none. The sections below unpack why.

\subsection{Model Size: Smaller is Better}\label{sec:results}

\begin{figure}[h]
    \centering
    \includegraphics[width=0.5\linewidth]{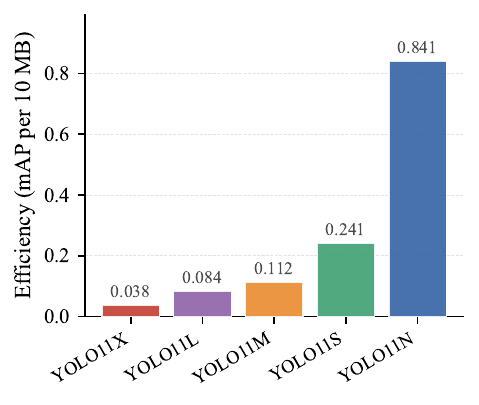}
    \caption{\textbf{Efficiency ranking of the evaluated models (mAP per 10\,MB).}
    YOLO11N achieves $22\times$ higher efficiency than YOLO11X while also reaching 
    the second highest absolute mAP$_{50}$ --- ruling out any accuracy--efficiency 
    tradeoff}
    \label{fig:main-fig}
\end{figure}

Figure~\ref{fig:main-fig} reports efficiency statistics across all YOLO11 variants. 
YOLO11N achieves the highest average efficiency (0.841) and the second highest 
absolute mAP$_{50}$ (0.459, versus 0.464 for YOLO11S --- a margin smaller than 
run-to-run variance) simultaneously (see Table~\ref{tab:efficiency} in 
Appendix~\ref{sec:additional-tables}), ruling out the hypothesis that larger models 
compensate their resource cost through superior detection performance. Its $21\times$ 
larger footprint makes YOLO11X's efficiency $22\times$ lower. Across all variants, 
the efficiency spread within a given model remains modest (11--19\%), confirming that 
architecture choice dominates over configuration choice. Robustness checks confirm 
this result across alternative efficiency measures (\cref{sec:additional-metrics}) 
and at fixed configurations (\cref{sec:robustness}).

\subsection{Why Small Models Outperform: The Data-Scarce Regime Penalizes Large 
Models}\label{sec:why}

On large benchmarks (COCO~\cite{lin2014microsoft}, DOTA~\cite{xia2018dota}, 
xView~\cite{lam2018xview}), larger models consistently dominate when data is 
abundant. We argue this intuition breaks down here, and provide two complementary 
pieces of evidence.

\paragraph{Large models depend on data volume they do not have.}

\begin{figure}[h]
    \centering
    \includegraphics[width=0.8\linewidth]{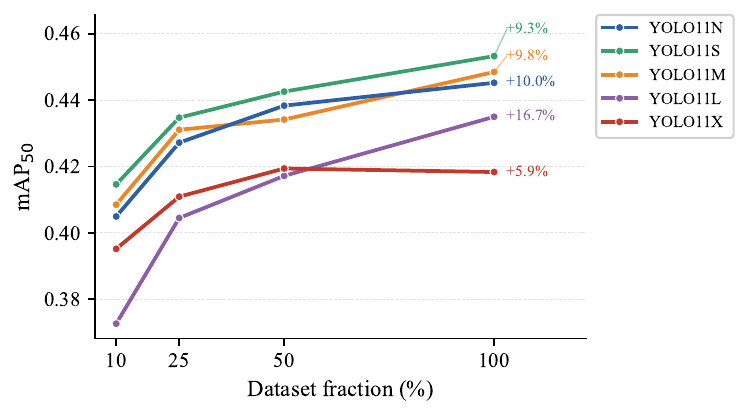}
    \caption{\textbf{Data efficiency curves for each YOLO11 variant averaged across 
    input resolutions.} Gains are front-loaded across all variants, with the largest 
    marginal improvements occurring between 10\% and 25\% of the training data. 
    YOLO11N saturates rapidly beyond 25\% ($+$0.007 mAP$_{50}$ from 50\% to 100\%), 
    while YOLO11L remains data-hungry throughout ($+$0.025 from 50\% to 100\%)}
    \label{fig:data_efficiency}
\end{figure}

Figure~\ref{fig:data_efficiency} shows data efficiency curves averaged across input 
resolutions. Performance gains are front-loaded across all variants: the largest 
marginal improvements occur between 10\% and 25\% of the training data, with 
diminishing returns beyond 50\%. The rate of saturation, however, varies with model 
size. YOLO11N exhibits the steepest early gain ($+$0.022 mAP$_{50}$ from 10\% to 
25\%) followed by rapid saturation, with only $+$0.007 additional gain from 50\% to 
100\%. YOLO11L presents the contrasting profile: gains remain substantial throughout 
($+$0.032 from 10\% to 25\%, $+$0.025 from 50\% to 100\%), confirming that larger 
models continue to benefit from additional data precisely where smaller ones have 
already saturated --- and where that data is unavailable in practice.

\paragraph{Overparameterization penalizes generalization.}

\begin{figure}[h]
    \centering
    \includegraphics[width=0.8\linewidth]{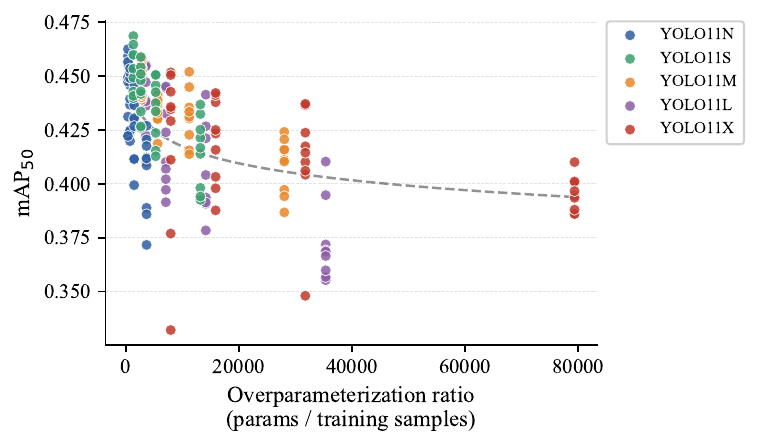}
    \caption{\textbf{mAP$_{50}$ vs.\ overparameterization ratio 
    $\rho = \text{params} / N_{\text{train}}$.} A negative log-linear trend 
    confirms that higher overparameterization predicts lower performance, 
    consistent with the bias-variance tradeoff. Each point is one configuration 
    (model $\times$ fraction $\times$ resolution)}
    \label{fig:overparameterization}
\end{figure}

Figure~\ref{fig:overparameterization} plots mAP$_{50}$ against the 
overparameterization ratio $\rho = \text{params} / N_{\text{train}}$, ranging from 
${\sim}300$ (YOLO11N, 100\% data) to ${\sim}80{,}000$ (YOLO11X, 10\% data). A 
negative log-linear trend emerges across the full range of configurations, consistent 
with the classical bias-variance tradeoff~(\cite{vapnik2013nature}): configurations 
with excess capacity relative to available data systematically underperform their 
lower-$\rho$ counterparts. This effect is compounded by the structural absence of 
COCO domain transfer --- zero-shot evaluation yields mAP$_{50} \leq 0.004$ across 
all configurations (\cref{sec:zero_shot}), confirming that pretrained representations 
provide no exploitable prior for this task and that larger pretrained models confer 
no initialization advantage.

\paragraph{Theoretical grounding.}
These observations instantiate three convergent predictions from the literature 
(detailed in Section~\ref{sec:related}). The bias-variance tradeoff~(\cite{vapnik2013nature}) 
predicts that generalization degrades with excess capacity --- our $\rho$ curve 
provides a direct empirical instantiation. The data-threshold crossover of 
\cite{brigato2021close} predicts smaller models should dominate below a sample 
count threshold --- our setting falls squarely in that regime. The Chinchilla 
scaling rule~(\cite{hoffmann2022training}) predicts that only YOLO11N operates near 
the compute-optimal token-to-parameter ratio for our training set; all larger 
variants are systematically undertrained relative to their capacity.

\subsection{Resource Allocation: Resolution or Data First?}\label{sec:resource_allocation}

Having established that model size should be minimized, we now ask: given a fixed 
budget, should practitioners prioritize more data or higher resolution?

Table~\ref{tab:efficiency_heatmap} reports average relative efficiency normalized by 
the baseline (416\,px, 10\%). Along the data axis at fixed resolution (416\,px), 
efficiency increases from 1.00 to 1.10 across the full annotation range --- a 10\% 
gain. Along the resolution axis at fixed data volume (10\%), moving from 416\,px to 
1280\,px yields a comparable 8\% gain. Data volume and resolution thus contribute 
similarly to efficiency when considered in isolation, and their effects are largely 
additive: the best configurations combine both high resolution and full data (1.17 
at 640\,px and 1280\,px).

\begin{table}[h]
\centering
\caption{\textbf{Average relative efficiency normalized by baseline (416\,px, 10\%).}
Best per resolution in \textbf{bold}. Along both axes, gains are modest and comparable: 
moving from 10\% to 100\% of the data at 416\,px yields $+$10\%, while moving from 
416\,px to 1280\,px at 10\% data yields $+$8\%}
\label{tab:efficiency_heatmap}
\begin{tabular}{lcccc}
\toprule
\textbf{Resolution} & \textbf{10\%} & \textbf{25\%} & \textbf{50\%} & \textbf{100\%} \\
\midrule
416\,px  & 1.00 & 1.06 & 1.08 & \textbf{1.10} \\
640\,px  & 1.06 & 1.11 & 1.12 & \textbf{1.17} \\
1280\,px & 1.08 & 1.15 & 1.14 & \textbf{1.17} \\
\bottomrule
\end{tabular}
\end{table}

Efficiency alone, however, does not determine the recommendation. The case for 
prioritizing resolution rests on two arguments that go beyond the efficiency table. 
First, resolution unlocks absolute detection performance that data volume alone 
cannot reach: YOLO11N at 1280\,px achieves mAP$_{50}$ = 0.459 versus 0.422 at 
416\,px with full data --- an 8.7\% gap that persists regardless of annotation 
budget (Table~\ref{tab:resolution_model_interaction} in 
\cref{sec:interaction-resolution}). Second, and more decisively, resolution is 
what drives Pareto dominance (Section~\ref{sec:pareto}): no amount of additional 
data at low resolution produces configurations that simultaneously maximize accuracy 
and throughput.

The mechanism is structural. Since COCO pretraining provides no exploitable prior 
(\cref{sec:zero_shot}), the model must learn discriminative features entirely from 
fine-tuning data. At 416\,px, aliasing artefacts destroy the fine-grained texture 
and edge detail that make PV panels identifiable --- features that cannot be 
recovered by adding more images at the same resolution. 

\begin{figure}[h]
    \centering
    \includegraphics[width=0.8\linewidth]{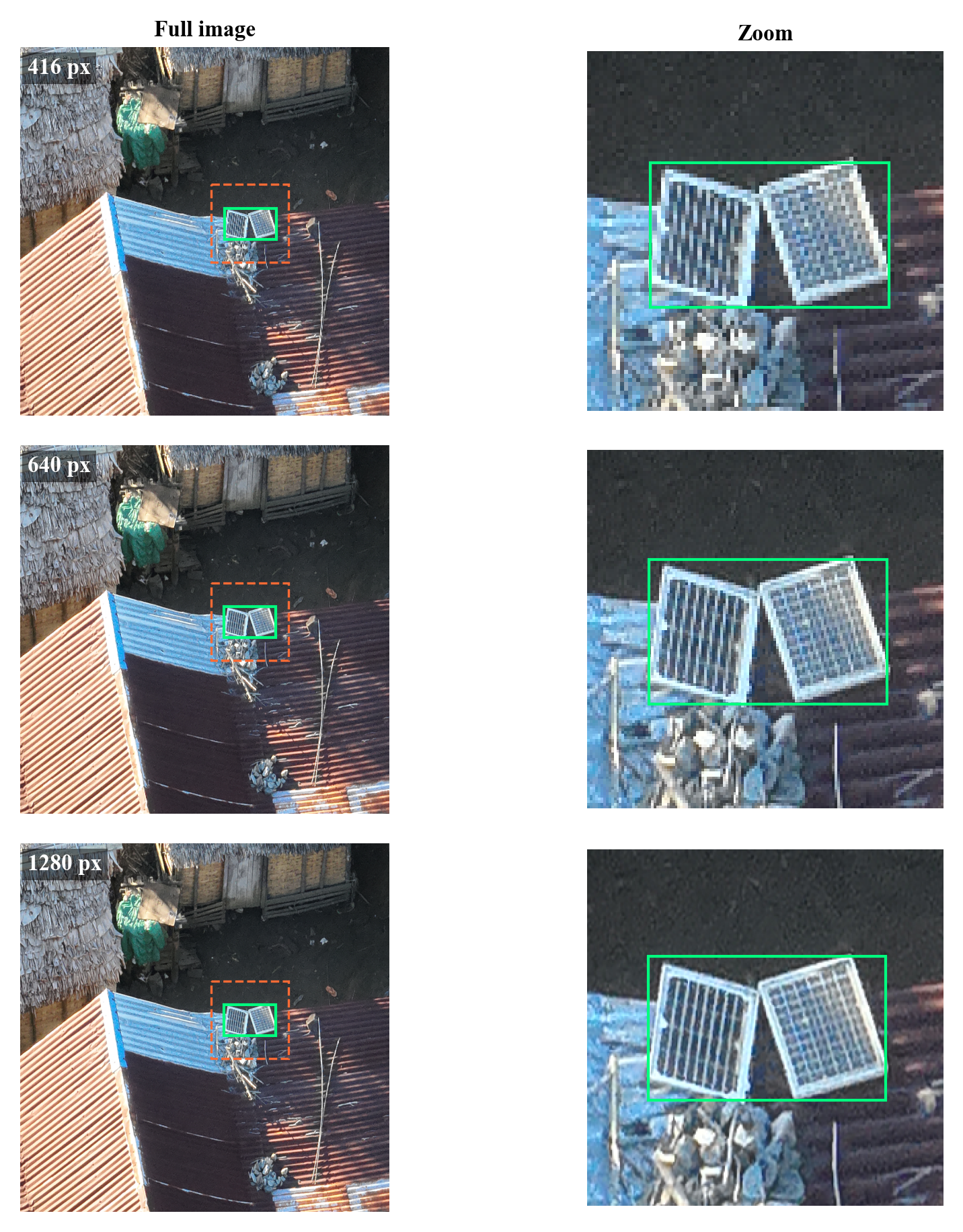}
    \caption{\textbf{Effect of input resolution on visual information for PV 
    detection.} Same scene at 416\,px (top), 640\,px (middle), 1280\,px (bottom). 
    Right column: zoomed crop at fixed physical size. At 416\,px the panel texture 
    is aliased and geometrically ambiguous; at 1280\,px cell structure and edges 
    are clearly resolved}
    \label{fig:resolution-effect}
\end{figure}

Higher resolutions progressively restore these high-frequency components (Figure~\ref{fig:resolution-effect}), consistent 
with~\cite{kasmi2025space}, who show that high-frequency features are 
disproportionately fragile under distribution shifts including GSD variations. 
Resolution is therefore not merely a performance lever but a prerequisite for 
accessing the discriminative signal in this task.

Finally, the efficiency ratio between YOLO11N and larger variants remains stable 
at ${\approx}3.5\times$ across all resolutions (\cref{sec:interaction-resolution}), 
confirming that resolution and model size operate on orthogonal axes: resolution 
scales efficiency proportionally across all model sizes without altering the 
relative ranking. The practitioner recommendation to select the smallest model 
holds independently of the resolution chosen.

\paragraph{Practical recommendation.} Prioritize resolution over data volume 
- not because resolution dominates efficiency, where both levers contribute 
comparably, but because resolution is the only lever that unlocks the absolute 
performance ceiling and drives Pareto dominance across all deployment objectives. 
Given a fixed budget: select YOLO11N, maximize input resolution, then invest 
remaining resources in additional annotations.
\section{Conclusion}
\subsection{Summary of Findings}

This paper investigated the efficiency of YOLO11 model variants across three resource
dimensions (model size, dataset fraction, and input resolution) for rooftop PV detection
in a data-scarce Earth observation setting. Three findings emerge consistently from our
analysis.

First, smaller models dominate and the data-scarce regime explains why. YOLO11N
achieves the second highest absolute mAP$_{50}$ (0.459, indistinguishable from
YOLO11S within run-to-run variance) and a $22\times$ efficiency advantage over
YOLO11X. This inversion is not accidental: data efficiency curves reveal that YOLO11N
reaches near-peak performance at 10\% of the training data, while the
overparameterization ratio $\rho = \text{params}/N_{\text{train}}$ negatively predicts
mAP$_{50}$ across all configurations --- consistent with the bias-variance tradeoff,
the Chinchilla compute-optimal regime~(\cite{hoffmann2022training}), and the complete
absence of exploitable COCO domain transfer.

Second, resolution and data volume contribute comparably to efficiency in isolation
(+10\% and +8\% respectively across their full ranges), but resolution is the
dominant resource allocation lever for two reasons that go beyond efficiency: it
unlocks absolute detection performance that additional data alone cannot reach
(+8.7\% mAP$_{50}$ from 416\,px to 1280\,px at fixed data), and it is the sole
driver of Pareto dominance in the joint accuracy--throughput space. The practical
recommendation follows: \textbf{select the smallest model, maximize input resolution,
then invest remaining budget in additional annotations.}

Third, these findings are robust to the choice of deployment objective.
Figure~\ref{fig:pareto-efficiency} confirms that small high-resolution configurations
are not only statically efficient but also operationally dominant: YOLO11N and
YOLO11S trained at 1280\,px lie at the apex of the Pareto frontier across all 60
configurations, simultaneously maximising detection accuracy and inference speed.
The same configurations that minimize model size and maximize resolution are optimal
regardless of whether the practitioner prioritizes static efficiency or real-time
deployment --- leaving no tradeoff to resolve.

More broadly, these results show that the scaling prior is not wrong but
\textit{conditional}: it holds when data abundance and domain transfer are both
satisfied, and inverts when neither is. In data-scarce EO settings, where both
conditions fail structurally, the default assumption that bigger is better should
be replaced by a principled efficiency analysis of the kind demonstrated here.

\subsection{Limitations}

Two conditions bound the scope of our conclusions. First, our recommendations are
established for the YOLO11 single-stage detector family; the efficiency inversion is
a property of this architecture class under data scarcity, and we do not claim it
transfers to two-stage detectors or transformer-based architectures without further
evidence. Second, our conclusions are drawn from a single geographic context and
object type (Madagascar rooftop PV); the mechanism we identify --- overparameterization
under annotation scarcity, absence of COCO domain transfer --- is general, but the
specific crossover thresholds we report are not guaranteed to hold for other EO
detection tasks without empirical validation.

\subsection{Future Work}

This paper establishes an efficiency inversion for a specific architecture family,
task, and geography; the natural next step is mapping where this inversion holds
and where it breaks. Three directions follow.

First, testing whether the inversion generalizes across architecture classes -
vision transformers, two-stage detectors, EfficientNet variants --- would clarify
whether it is a structural property of the data-scarce regime or specific to
single-stage detectors, and would allow formalizing efficiency-based scaling laws
for resource-constrained EO more broadly. This includes comparing against methods
purpose-built for small object detection, such as super-resolution preprocessing or
scale-aware training~(\cite{shermeyer2019effects,singh2018sniper}), which may
interact with model size and resolution scaling in ways our factorial design did
not isolate, and extending the resolution range beyond 1280\,px, where GPU memory
constrained our current grid.

Second, real-world deployment studies measuring operational efficiency on edge
devices would close the loop between our training-time analysis and the inference
constraints practitioners actually face in developing countries.

Third, evaluating domain-adaptive or self-supervised pretraining on unlabeled EO
imagery would test whether a stronger initialization shifts the crossover point
between model sizes --- our result is explained in part by the complete absence of
exploitable COCO transfer, and a better prior could alter the relative advantage
of small models we observe here.

More broadly, extending this framework to other data-scarce EO tasks --- crop type
mapping, building footprint extraction, damage assessment --- and to broader
geographic coverage than our current city-level split allows would establish how
far the resource allocation recommendation travels beyond PV detection in
Madagascar.

\begin{Backmatter}

\section{Acknowledgements}

\paragraph{Acknowledgments} This work originates from a broader applied statistics project. We would like to sincerely thank Malo David for his substantial work on data processing and curation, which laid the foundations for the present study. We are also grateful to Maxime Chansat and Thomas Lambelin for insightful discussions and their contributions to the initial stages of the project. Their input and collaboration were instrumental in shaping the direction from which this paper emerged.


\paragraph{Competing Interests}
The authors declare none.

\paragraph{Data Availability Statement}

The solar panel dataset used in this article is publicly available on \href{https://zenodo.org/records/15336414}{Zenodo}. The code used for model training, inference, and evaluation is publicly available on \href{https://github.com/kwame-mbobda-kuate/scaling-laws-eds}{GitHub}.

\paragraph{Ethical Standards}
The research meets all ethical guidelines, including adherence to the legal requirements of the study country.

\paragraph{Author Contributions}
Conceptualization: G.K.; Data curation: K.M.; Formal analysis: K.M.; 
Investigation: K.M.; Methodology: K.M., G.K.; Validation: K.M., G.K.; 
Visualization: K.M.; Writing -- original draft: K.M.; 
Writing -- review \& editing: K.M., G.K.; Funding acquisition: N/A; 
Resources: K.M.; Supervision: G.K.; Project administration: G.K. 
All authors approved the final submitted draft.


\printbibliography

@String(CVPR= {IEEE Conf. Comput. Vis. Pattern Recog.})

@String(ICPR = {Int. Conf. Pattern Recog.})

@String(CVPR  = {CVPR})

@String(ICPR  = {ICPR})

@phdthesis{kasmi_enhancing_2024,
	type = {Theses},
	title = {Enhancing the {Reliability} of {Deep} {Learning} {Models} to {Improve} the {Observability} of {French} {Rooftop} {Photovoltaic} {Installations}},
	copyright = {All rights reserved},
	school = {Université Paris sciences et lettres},
	author = {Kasmi, Gabriel},
	month = apr,
	year = {2024},
	note = {Issue: 2024UPSLM027},
}

@misc{OpenStat,
  url = {https://zenodo.org/records/15120875},
  year = {2025},
  publisher = {OpenStat Madagascar},
  author = {{OpenStat Madagascar}},
  title = {{Données sur l’énergie solaire et labellisation d’images de panneaux photovoltaïques à Madagascar}}
}

@article{kaplan2020scaling,
  title={{Scaling laws for neural language models}},
  author={Kaplan, Jared and McCandlish, Sam and Henighan, Tom and Brown, Tom B and Chess, Benjamin and Child, Rewon and Gray, Scott and Radford, Alec and Wu, Jeffrey and Amodei, Dario},
  journal={arXiv preprint arXiv:2001.08361},
  year={2020}
}

@inproceedings{hoffmann2022training,
 author = {Hoffmann, Jordan and Borgeaud, Sebastian and Mensch, Arthur and Buchatskaya, Elena and Cai, Trevor and Rutherford, Eliza and de Las Casas, Diego and Hendricks, Lisa Anne and Welbl, Johannes and Clark, Aidan and Hennigan, Thomas and Noland, Eric and Millican, Katherine and van den Driessche, George and Damoc, Bogdan and Guy, Aurelia and Osindero, Simon and Simonyan, Kar\'{e}n and Elsen, Erich and Vinyals, Oriol and Rae, Jack and Sifre, Laurent},
 booktitle = {Advances in Neural Information Processing Systems},
 editor = {S. Koyejo and S. Mohamed and A. Agarwal and D. Belgrave and K. Cho and A. Oh},
 pages = {30016--30030},
 publisher = {Curran Associates, Inc.},
 title = {{An empirical analysis of compute-optimal large language model training}},
 volume = {35},
 year = {2022}
}

@article{cheng2020remote,
  title={{Remote sensing image scene classification: Benchmark and state of the art}},
  author={Cheng, Gong and Xie, Xingxing and Han, Junwei and Guo, Lei and Xia, Gui-Song},
  journal={Proceedings of the IEEE},
  volume={105},
  number={10},
  pages={1865--1883},
  year={2017},
  publisher={IEEE}
}

@software{yolo11_ultralytics,
  author = {Glenn Jocher and Jing Qiu},
  title = {{Ultralytics YOLO11}},
  version = {11.0.0},
  year = {2024},
  url = {https://github.com/ultralytics/ultralytics},
  license = {AGPL-3.0}
}

@report{iea2023world,
  title={{World Energy Outlook 2023}},
  author={{International Energy Agency}},
  year={2023},
  institution={IEA Publications}
}

@inproceedings{tan2019efficientnet,
  title={{Efficientnet: Rethinking model scaling for convolutional neural networks}},
  author={Tan, Mingxing and Le, Quoc},
  booktitle={International conference on machine learning},
  pages={6105--6114},
  year={2019},
  organization={PMLR}
}

@inproceedings{
frankle2018lottery,
title={{The Lottery Ticket Hypothesis: Finding Sparse, Trainable Neural Networks}},
author={Jonathan Frankle and Michael Carbin},
booktitle={International Conference on Learning Representations},
year={2019},
}

@inproceedings{
dosovitskiy2021image,
title={{An Image is Worth 16x16 Words: Transformers for Image Recognition at Scale}},
author={Alexey Dosovitskiy and Lucas Beyer and Alexander Kolesnikov and Dirk Weissenborn and Xiaohua Zhai and Thomas Unterthiner and Mostafa Dehghani and Matthias Minderer and Georg Heigold and Sylvain Gelly and Jakob Uszkoreit and Neil Houlsby},
booktitle={International Conference on Learning Representations},
year={2021},
}

@InProceedings{zhai2022scaling,
    author    = {Zhai, Xiaohua and Kolesnikov, Alexander and Houlsby, Neil and Beyer, Lucas},
    title     = {{Scaling Vision Transformers}},
    booktitle = {Proceedings of the IEEE/CVF Conference on Computer Vision and Pattern Recognition (CVPR)},
    month     = {June},
    year      = {2022},
    pages     = {12104-12113}
}

@article{yu2018solar,
title = {{DeepSolar: A Machine Learning Framework to Efficiently Construct a Solar Deployment Database in the United States}},
journal = {Joule},
volume = {2},
number = {12},
pages = {2605-2617},
year = {2018},
issn = {2542-4351},
author = {Jiafan Yu and Zhecheng Wang and Arun Majumdar and Ram Rajagopal}
}

@article{kasmi2023crowdsourced,
   title={{A crowdsourced dataset of aerial images with annotated solar photovoltaic arrays and installation metadata}},
   volume={10},
   ISSN={2052-4463},
   number={1},
   journal={Scientific Data},
   publisher={Springer Science and Business Media LLC},
   author={Kasmi, Gabriel and Saint-Drenan, Yves-Marie and Trebosc, David and Jolivet, Raphaël and Leloux, Jonathan and Sarr, Babacar and Dubus, Laurent},
   year={2023},
   month=jan }

@misc{beyer2022better,
      title={{Better plain ViT baselines for ImageNet-1k}}, 
      author={Lucas Beyer and Xiaohua Zhai and Alexander Kolesnikov},
      year={2022},
      eprint={2205.01580},
      archivePrefix={arXiv},
      primaryClass={cs.CV},
}

@article{singh2018analysis,
  title={{An Analysis of Scale Invariance in Object Detection - SNIP}},
  author={Bharat Singh and Larry S. Davis},
  journal={2018 IEEE/CVF Conference on Computer Vision and Pattern Recognition},
  year={2017},
  pages={3578-3587},
}

@inproceedings{singh2018sniper,
 author = {Singh, Bharat and Najibi, Mahyar and Davis, Larry S},
 booktitle = {Advances in Neural Information Processing Systems},
 editor = {S. Bengio and H. Wallach and H. Larochelle and K. Grauman and N. Cesa-Bianchi and R. Garnett},
 pages = {},
 publisher = {Curran Associates, Inc.},
 title = {{SNIPER: Efficient Multi-Scale Training}},
 volume = {31},
 year = {2018}
}

@InProceedings{lin2017feature,
author = {Lin, Tsung-Yi and Dollar, Piotr and Girshick, Ross and He, Kaiming and Hariharan, Bharath and Belongie, Serge},
title = {{Feature Pyramid Networks for Object Detection}},
booktitle = {Proceedings of the IEEE Conference on Computer Vision and Pattern Recognition (CVPR)},
month = {July},
year = {2017}
}

@article{cheng2016survey,
   title={{A survey on object detection in optical remote sensing images}},
   volume={117},
   ISSN={0924-2716},
   journal={ISPRS Journal of Photogrammetry and Remote Sensing},
   publisher={Elsevier BV},
   author={Cheng, Gong and Han, Junwei},
   year={2016},
   month=jul, pages={11–28} 
   }

@inproceedings{dvornik2018modeling,
  title={{Modeling Visual Context is Key to Augmenting Object Detection Datasets}},
  author={Nikita Dvornik and Julien Mairal and Cordelia Schmid},
  booktitle={European Conference on Computer Vision},
  year={2018}
}

@inproceedings{snell2017prototypical,
  title={{Prototypical Networks for Few-shot Learning}},
  author={Jake Snell and Kevin Swersky and Richard S. Zemel},
  booktitle={Neural Information Processing Systems},
  year={2017},
}

@inproceedings{yosinski2014transferable,
 author = {Yosinski, Jason and Clune, Jeff and Bengio, Yoshua and Lipson, Hod},
 booktitle = {Advances in Neural Information Processing Systems},
 editor = {Z. Ghahramani and M. Welling and C. Cortes and N. Lawrence and K.Q. Weinberger},
 pages = {},
 publisher = {Curran Associates, Inc.},
 title = {{How transferable are features in deep neural networks?}},
 volume = {27},
 year = {2014}
}

@InProceedings{finn2017maml,
  title = 	 {{Model-Agnostic Meta-Learning for Fast Adaptation of Deep Networks}},
  author =       {Chelsea Finn and Pieter Abbeel and Sergey Levine},
  booktitle = 	 {Proceedings of the 34th International Conference on Machine Learning},
  pages = 	 {1126--1135},
  year = 	 {2017},
  editor = 	 {Precup, Doina and Teh, Yee Whye},
  volume = 	 {70},
  series = 	 {Proceedings of Machine Learning Research},
  month = 	 {06--11 Aug},
  publisher =    {PMLR},
}

@inproceedings{parhar2022hyperionsolarnet,
  title={{HyperionSolarNet: Solar Panel Detection from Aerial Images}},
  author={Parhar, Poonam and Sawasaki, Ryan and Todeschini, Alberto and Reed, Colorado and Vahabi, Hossein and Nusaputra, Nathan and Vergara, Felipe},
  booktitle={NeurIPS 2021 Workshop on Tackling Climate Change with Machine Learning},
  year={2021}
}

@article{Mayer2022_3DPVLocator,
  title = {{3D-PV-Locator: Large-scale Detection of Rooftop-mounted Photovoltaic Systems in 3D}},
  author = {Mayer, Kevin and Rausch, Benjamin and Arlt, Marie-Louise and Gust, Gunther and Wang, Zhecheng and Neumann, Dirk and Rajagopal, Ram},
  journal = {Applied Energy},
  volume = {310},
  pages = {118469},
  year = {2022},
}

@article{Kausika2021_GeoAI_Netherlands,
  title = {{GeoAI for Detection of Solar Photovoltaic Installations in the Netherlands}},
  author = {Kausika, Bala Bhavya and Nijmeijer, Diede and Reimerink, Iris and Brouwer, Peter and Liem, Vera},
  journal = {Energy and AI},
  volume = {6},
  pages = {100111},
  year = {2021},
}

@inproceedings{Kasmi2022_TowardsUnsupervisedAssessment,
  author = {Kasmi, Gabriel and Dubus, Laurent and Blanc, Philippe and Saint-Drenan, Yves-Marie},
  title = {{Towards Unsupervised Assessment with Open‑Source Data of the Accuracy of Deep Learning‑Based Distributed PV Mapping}},
  booktitle = {Proceedings of the Workshop on Machine Learning for Earth Observation (MACLEAN)},
  year = {2022},
  note = {in conjunction with ECML\/PKDD 2022},
}

@article{bouaziz2024high,
  title={{High-resolution solar panel detection in Sfax, Tunisia: A UNet-Based approach}},
  author={Bouaziz, Mohamed Chahine and El Koundi, Mourad and Ennine, Ghaleb},
  journal={Renewable Energy},
  volume={235},
  pages={121171},
  year={2024},
  publisher={Elsevier}
}

@book{vapnik2013nature,
  title={The nature of statistical learning theory},
  author={Vapnik, Vladimir},
  year={2013},
  publisher={Springer science \& business media}
}

@inproceedings{brigato2021close,
  title={A close look at deep learning with small data},
  author={Brigato, Lorenzo and Iocchi, Luca},
  booktitle={2020 25th international conference on pattern recognition (ICPR)},
  pages={2490--2497},
  year={2021},
  organization={IEEE}
}

@inproceedings{
shi2024when,
title={{When Do We Not Need Larger Vision Models?}},
author={Baifeng Shi and Ziyang Wu and Maolin Mao and Xin Wang and Trevor Darrell},
booktitle={NeurIPS 2024 Workshop: Self-Supervised Learning - Theory and Practice},
year={2024},
}

@inproceedings{pardo2021baod,
  title={{BAOD: budget-aware object detection}},
  author={Pardo, Alejandro and Xu, Mengmeng and Thabet, Ali and Arbel{\'a}ez, Pablo and Ghanem, Bernard},
  booktitle={Proceedings of the IEEE/CVF Conference on Computer Vision and Pattern Recognition},
  pages={1247--1256},
  year={2021}
}

@inproceedings{shermeyer2019effects,
  title={The effects of super-resolution on object detection performance in satellite imagery},
  author={Shermeyer, Jacob and Van Etten, Adam},
  booktitle={Proceedings of the IEEE/CVF Conference on Computer Vision and Pattern Recognition Workshops},
  pages={0--0},
  year={2019}
}

@article{bradbury2016distributed,
  title={Distributed solar photovoltaic array location and extent dataset for remote sensing object identification},
  author={Bradbury, Kyle and Saboo, Raghav and L Johnson, Timothy and Malof, Jordan M and Devarajan, Arjun and Zhang, Wuming and M Collins, Leslie and G Newell, Richard},
  journal={Scientific data},
  volume={3},
  number={1},
  pages={160106},
  year={2016},
  publisher={Nature Publishing Group}
}

@article{furedi2026labeled,
  title={{Labeled photovoltaic installations for orthographic aerial imagery in Queens, New York}},
  author={Furedi, Tyler and Kimsal, Edwin and Cornejo, Samara and Liero, Nicholas and Ranalli, Joseph},
  journal={Scientific Data},
  year={2026},
  publisher={Nature Publishing Group UK London}
}

@article{kruitwagen2021global,
  title={A global inventory of photovoltaic solar energy generating units},
  author={Kruitwagen, Lucas and Story, KT and Friedrich, J and Byers, L and Skillman, S and Hepburn, Cameron},
  journal={Nature},
  volume={598},
  number={7882},
  pages={604--610},
  year={2021},
  publisher={Nature Publishing Group UK London}
}

@article{li2025global,
  title={Global photovoltaic solar panel dataset from 2019 to 2022},
  author={Li, Anqi and Liu, Luling and Li, Shijie and Cui, Xihong and Chen, Xuehong and Cao, Xin},
  journal={Scientific Data},
  volume={12},
  number={1},
  pages={637},
  year={2025},
  publisher={Nature Publishing Group UK London}
}

@article{kasmi2025space,
  title={Space-scale exploration of the poor reliability of deep learning models: the case of the remote sensing of rooftop photovoltaic systems},
  author={Kasmi, Gabriel and Dubus, Laurent and Saint-Drenan, Yves-Marie and Blanc, Philippe},
  journal={Environmental Data Science},
  volume={4},
  pages={e22},
  year={2025},
  publisher={Cambridge University Press}
}

@inproceedings{lin2014microsoft,
  title={{Microsoft COCO: Common Objects in Context}},
  author={Lin, Tsung-Yi and Maire, Michael and Belongie, Serge and Hays, James and Perona, Pietro and Ramanan, Deva and Doll{\'a}r, Piotr and Zitnick, C Lawrence},
  booktitle={European conference on computer vision},
  pages={740--755},
  year={2014},
  organization={Springer}
}

@inproceedings{xia2018dota,
  title={{DOTA: A large-scale dataset for object detection in aerial images}},
  author={Xia, Gui-Song and Bai, Xiang and Ding, Jian and Zhu, Zhen and Belongie, Serge and Luo, Jiebo and Datcu, Mihai and Pelillo, Marcello and Zhang, Liangpei},
  booktitle={Proceedings of the IEEE conference on computer vision and pattern recognition},
  pages={3974--3983},
  year={2018}
}

@misc{lam2018xview,
  title         = {{xView: Objects in Context in Overhead Imagery}},
  author        = {Lam, Darius and Kuzma, Richard and McGee, Kevin and 
                   Dooley, Samuel and Laielli, Michael and Klaric, Matthew and 
                   Bulatov, Yaroslav and McCord, Brendan},
  year          = {2018},
  eprint        = {1802.07856},
  archivePrefix = {arXiv},
  primaryClass  = {cs.CV}
}

@article{rustowicz2019semantic,
  title={Semantic segmentation of crop type in Africa: A novel dataset and analysis of deep learning methods},
  author={Rustowicz, Rose and Cheong, Robin and Wang, Lijing and Ermon, Stefano and Burke, Marshall and Lobell, David},
  journal={Proceedings of the IEEE/CVF Conference on Computer Vision and Pattern Recognition Workshops},
  year={2019}
}

\begin{appendix}
\clearpage
\setcounter{page}{1}

\section{Additional details on the dataset}\label{sec:additional_dataset}

\paragraph{Geographic distribution}
Figure~\ref{fig:localization} presents the geographic distribution of the 8,977 drone 
images retained for our analysis, colored by split assignment. Images are concentrated 
along major transportation corridors and around Antananarivo, reflecting the crowdsourced 
collection methodology, with secondary clusters along the east coast and in the southern 
regions around Fianarantsoa and Toliara. The random split yields a spatially representative 
test set covering the full geographic extent of the dataset.

\begin{figure}[h]
    \centering
    \includegraphics[width=.6\linewidth]{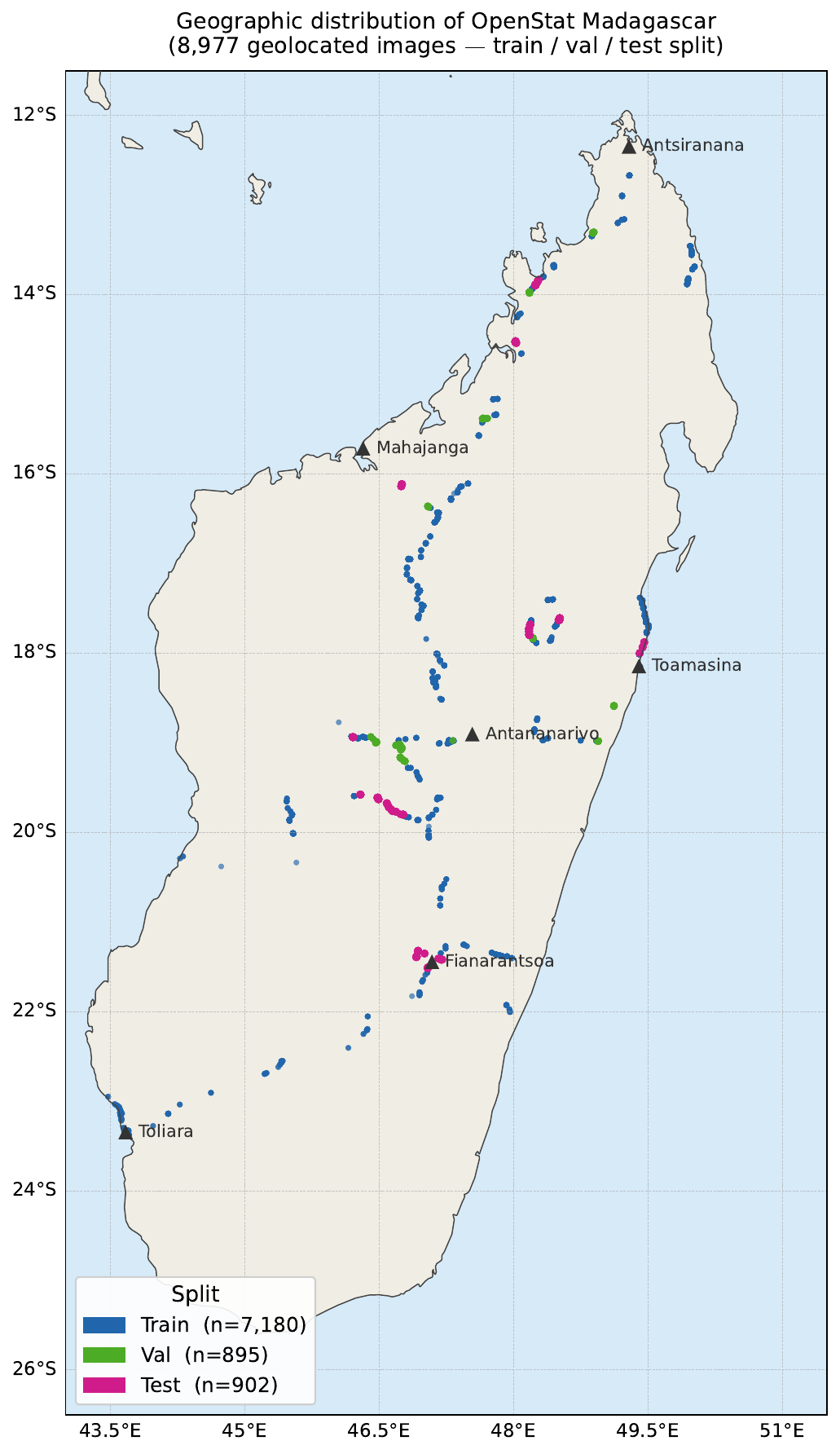}
    \caption{\textbf{Geographic distribution of the 8,977 drone images retained for 
    training, colored by split assignment (train / val / test).} Sampling points follow 
    major transportation corridors and population centers, with Antananarivo as the 
    primary cluster. The random split preserves the spatial coverage of the full dataset 
    across all three subsets}
    \label{fig:localization}
\end{figure}

\paragraph{Representative images}
Figure~\ref{fig:drones_examples} shows two representative drone images from the OpenStat 
Madagascar dataset, illustrating the diversity of rooftop configurations, installation 
types, and imaging conditions. Individual solar panels are clearly visible but occupy a 
small fraction of the total image area, consistent with the bounding box size distribution 
reported in Figure~\ref{fig:bbox_cdf}.

\begin{figure}[htb]
    \centering
    \includegraphics[width=0.4\textwidth]{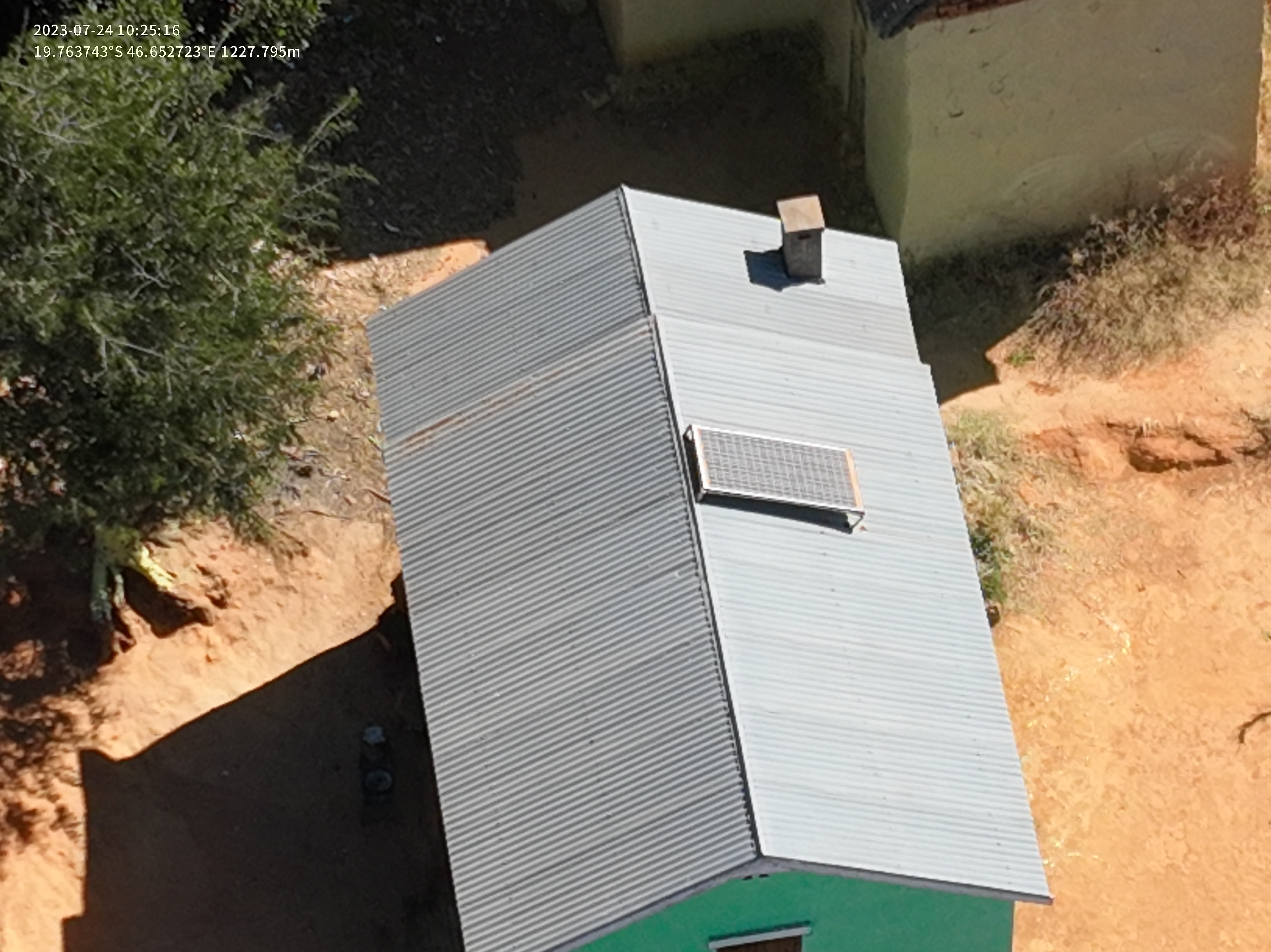}
    \hfill
    \includegraphics[width=0.4\textwidth]{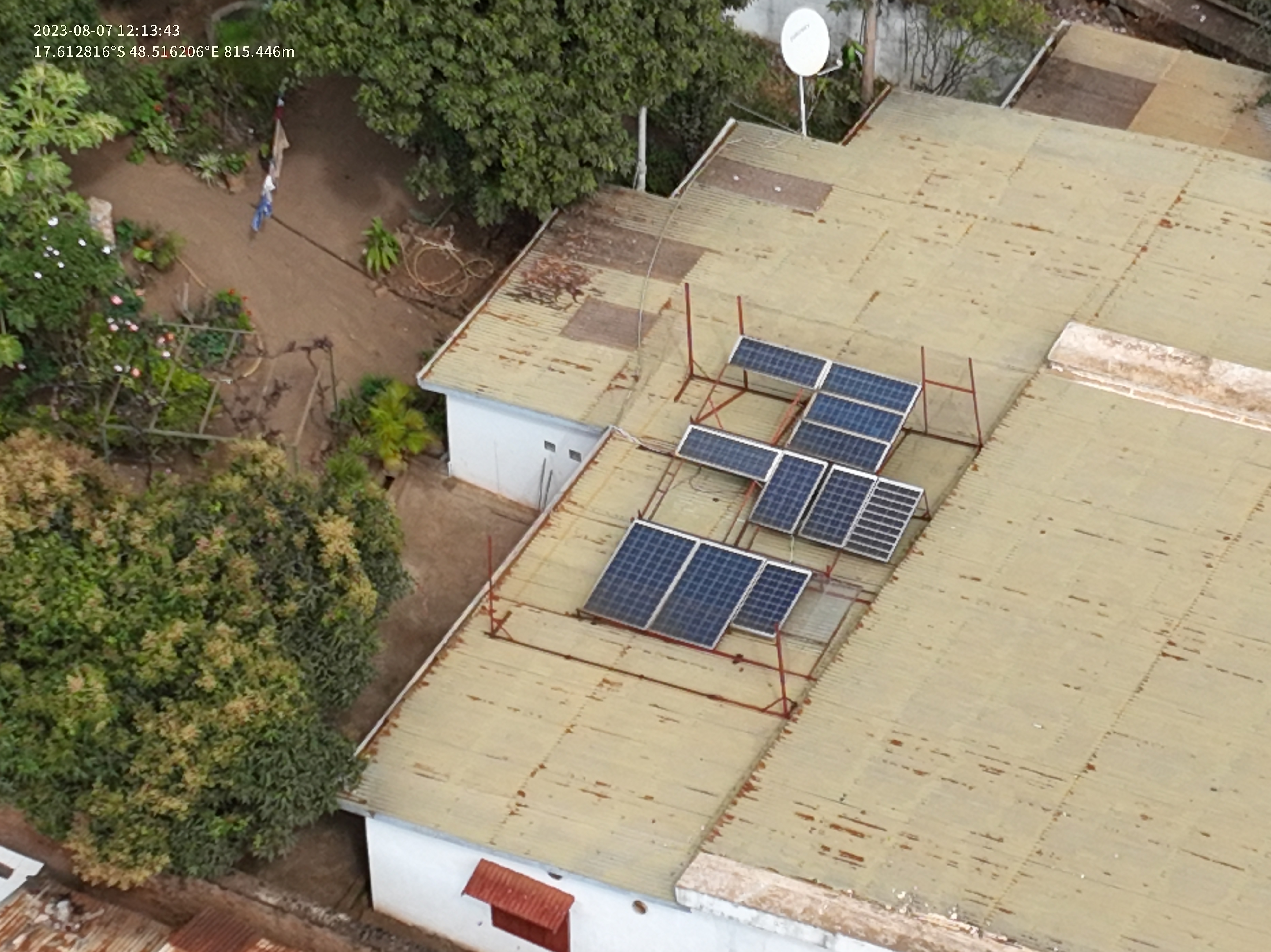}
    \caption{\textbf{Representative high-resolution drone images from the OpenStat 
    Madagascar dataset.} Images illustrate the variety of rooftop types, 
    panel orientations, and background clutter encountered across acquisition sites}
    \label{fig:drones_examples}
\end{figure}

\paragraph{Annotated examples}
Figure~\ref{fig:bbox_examples} shows examples of images with ground-truth bounding box 
annotations from the validation set. Each box corresponds to an individual solar panel or 
boiler identified by a human annotator. The high density of small, tightly-packed boxes 
in some images illustrates the detection difficulty and confirms the small object 
characterization established in Section~\ref{sec:data}.

\begin{figure}[h]
    \centering
    \includegraphics[width=0.8\linewidth]{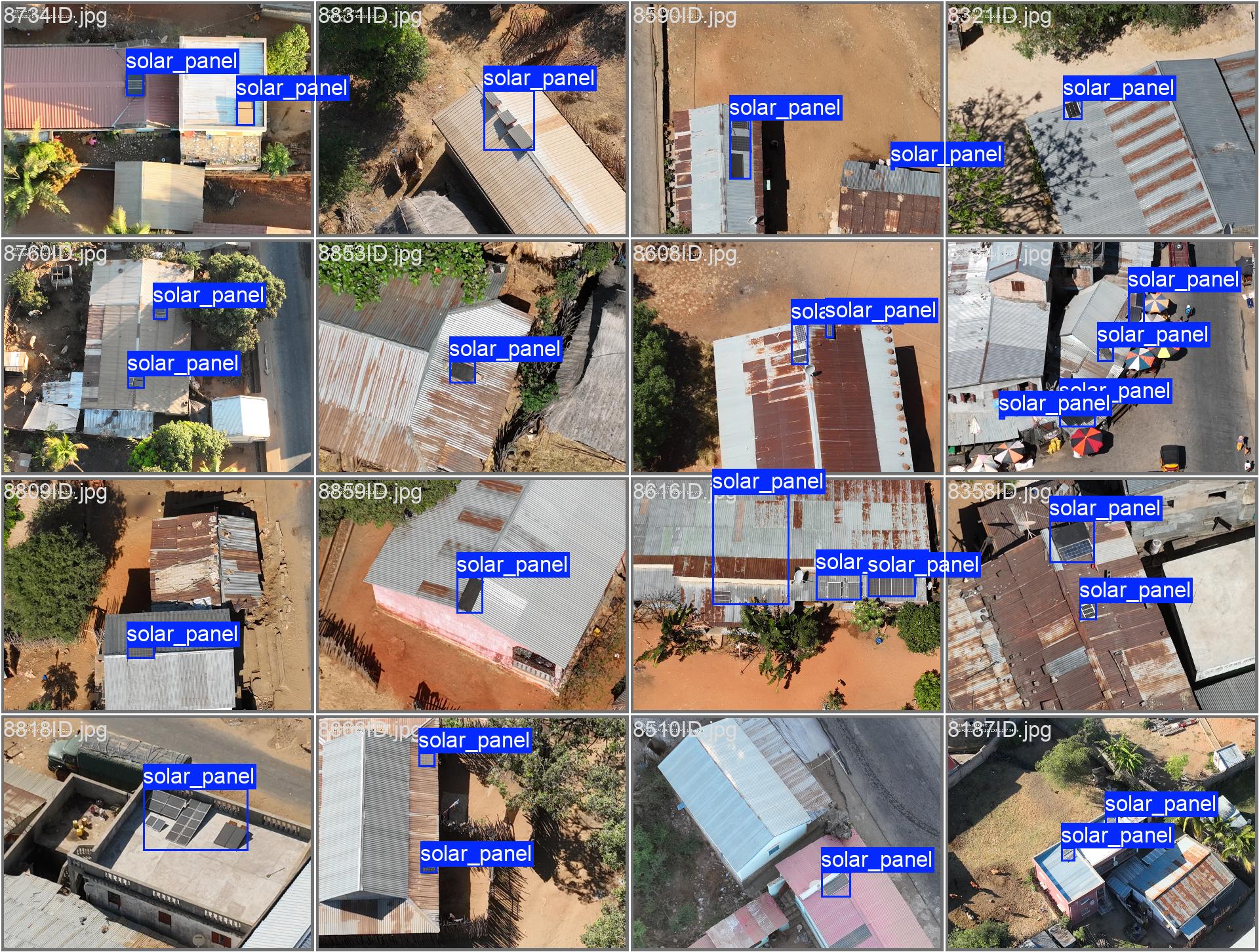}
    \caption{\textbf{Validation set images with ground-truth bounding box annotations.} 
    Dense clusters of small boxes are characteristic of rooftop installations with 
    multiple panels, and represent the dominant challenge in this detection task}
    \label{fig:bbox_examples}
\end{figure}

\newpage
\section{Additional experimental details}\label{sec:implementational-details}

This section will describe the implementational details --- seeds, learning rate, 
data augmentation, etc. The parameters mostly correspond to the default values 
set by Ultralytics and described in the 
\href{https://docs.ultralytics.com/modes/train#augmentation-settings-and-hyperparameters}{documentation}. 
Please check it for further information.

\begin{table}[h]
\centering
\caption{\textbf{Hyperparemeters used during the training.} The batch size is modulated to avoid out of memory errors depending on the resolution and the number of parameters of the model in question. This variability is compensated by gradient accumulation, which always simulated a (nominal) batch size of 64}
\begin{tabular}{lc}
\toprule
\textbf{Hyperparameter} & \textbf{Value} \\
\midrule
Batch size & \{2, 4, 8, 16, 32, 64\} \\
Number of epochs & 50 \\
Optimizer & Adam \\
Random seed & \{693763239, 3810243382, 4221376603\} \\
Initial learning rate & 0.001 \\
Final learning rate & 0.001 \\
Momentum & 0.937 \\
Weight decay & 0.0005 \\
Warmup epochs & 3.0 \\
Warmup momentum & 0.8 \\
Warmup learning rate of bias & 0.1 \\
Weight of box loss & 7.5 \\
Nominal batch size & 64 \\
\bottomrule
\end{tabular}%
\end{table}

\begin{table}[h]
\centering
\caption{\textbf{Data augmentations performed during training.} The mosaic is disabled on the last 10 epochs (41‑50)}
\begin{tabular}{lcc}
\toprule
\textbf{Augmentation} & \textbf{Scale} & \textbf{Probability} \\
\midrule
Hue variation & 0.015 & --- \\
Saturation variation & 0.7 & --- \\
Brightness variation & 0.4 & --- \\
Translation & 0.1 & --- \\
Rescaling & 0.5 & --- \\
Left-to-right flip & --- & 0.5 \\
Mosaic & --- & 1 \\
\bottomrule
\end{tabular}%
\end{table}

\newpage
\section{Additional tables}\label{sec:additional-tables}

\paragraph{Efficiency statistics}
Table~\ref{tab:efficiency} reports per-model efficiency statistics (min, max, average) 
across all completed training configurations, along with the best-performing 
(dataset fraction, resolution) pair and its corresponding mAP$_{50}$. YOLO11N 
dominates on all three statistics; notably, even its minimum efficiency (0.729) 
exceeds the maximum efficiency of any other variant.

\begin{table}[h]
\centering
\caption{\textbf{Efficiency statistics per model variant.} Efficiency is defined as mAP$_{50}$ / ModelSize~(MB). Best Config reports the (dataset fraction, resolution) achieving the highest efficiency, along with the corresponding mAP$_{50}$}
\begin{tabular}{lcccccc}
\toprule
\textbf{Model} & \textbf{Min Eff.} & \textbf{Max Eff.} & \textbf{Avg Eff.} & \textbf{Best Config} & \textbf{mAP$_{50}$} \\
\midrule
YOLO11N & 0.729 & 0.907 & 0.841 & 100\%~/~1280px & 0.463 \\
YOLO11S & 0.217 & 0.259 & 0.241 & 100\%~/~1280px & 0.469 \\
YOLO11M & 0.101 & 0.119 & 0.112 & 100\%~/~1280px & 0.458 \\
YOLO11L & 0.073 & 0.094 & 0.084 & 100\%~/~640px & 0.455 \\
YOLO11X & 0.031 & 0.042 & 0.038 & 100\%~/~640px & 0.452 \\
\bottomrule
\end{tabular}%
\label{tab:efficiency}
\end{table}

\section{Additional results}\label{sec:additional}

\subsection{Consistency Across Efficiency Measures}\label{sec:additional-metrics}

Table~\ref{tab:efficiency_variants} reports average efficiency across all training 
configurations for each model variant, computed under five alternative performance 
metrics. YOLO11N achieves the highest efficiency under all measures, and the ranking 
N $>$ S $>$ M $>$ L $>$ X is perfectly monotonic across all five columns --- no 
metric reversal is observed anywhere in the table. The efficiency gap between YOLO11N 
and YOLO11X ranges from $20\times$ (mAP$_{50}$ and mAP$_{50\text{-}95}$) to 
$23\times$ (Precision), confirming that the efficiency inversion reported in 
\cref{sec:results} is not an artifact of the choice of performance measure but 
a structural property of the data-scarce regime.

\begin{table}[h]
\centering
\caption{\textbf{Efficiency under alternative performance metrics.} Each column reports the average efficiency (metric / model size in MB $\times$ 10) across all training configurations for each model variant. YOLO11N consistently achieves the highest efficiency regardless of the metric used, confirming that the efficiency inversion reported in the main text is robust to the choice of performance measure}
\label{tab:efficiency_variants}
\begin{tabular}{lccccc}
\toprule
\textbf{Model} & \textbf{Eff. (mAP$_{50}$)} & \textbf{Eff. (mAP$_{50\text{-}95}$)} & \textbf{Eff. (Precision)} & \textbf{Eff. (Recall)} & \textbf{Eff. (F$_1$)} \\
\midrule
YOLO11N & 0.841 & 0.606 & 1.183 & 1.022 & 1.096 \\
YOLO11S & 0.241 & 0.175 & 0.332 & 0.293 & 0.311 \\
YOLO11M & 0.112 & 0.081 & 0.155 & 0.138 & 0.146 \\
YOLO11L & 0.084 & 0.060 & 0.118 & 0.103 & 0.109 \\
YOLO11X & 0.038 & 0.027 & 0.053 & 0.047 & 0.050 \\
\bottomrule
\end{tabular}%
\end{table}

\subsection{Robustness Across Configurations}\label{sec:robustness}

Table~\ref{tab:iso_config} confirms the dominance of YOLO11N under a fair, 
iso-configuration comparison. Restricting the analysis to 416\,px --- the only 
resolution at which all five variants were evaluated --- YOLO11N achieves the 
highest efficiency across all dataset fractions (1.056--1.109), with scores 
remaining stable regardless of data volume. This rules out the hypothesis that 
larger models might dominate under specific data regimes: even at 10\% of the 
dataset, YOLO11N outperforms all larger variants while maintaining competitive 
absolute mAP$_{50}$ (0.548).

\begin{table}[h]
\centering
\caption{\textbf{Best model per iso-configuration.} Only configurations where all five YOLO11 variants were evaluated are included. YOLO11N dominates across all tested configurations}
\label{tab:iso_config}
\begin{tabular}{cccccc}
\toprule
\textbf{Fraction} & \textbf{Resolution} & \textbf{Best Model} & \textbf{Efficiency} & \textbf{mAP$_{50}$} \\
\midrule
10\% & 416px & YOLO11N & 0.763 & 0.389 \\
25\% & 416px & YOLO11N & 0.807 & 0.411 \\
50\% & 416px & YOLO11N & 0.831 & 0.424 \\
100\% & 416px & YOLO11N & 0.828 & 0.422 \\
10\% & 640px & YOLO11N & 0.808 & 0.412 \\
25\% & 640px & YOLO11N & 0.843 & 0.430 \\
50\% & 640px & YOLO11N & 0.856 & 0.437 \\
100\% & 640px & YOLO11N & 0.893 & 0.456 \\
10\% & 1280px & YOLO11N & 0.825 & 0.421 \\
25\% & 1280px & YOLO11N & 0.876 & 0.447 \\
50\% & 1280px & YOLO11N & 0.881 & 0.449 \\
100\% & 1280px & YOLO11N & 0.901 & 0.459 \\
\bottomrule
\end{tabular}%
\end{table}

\subsection{Zero-Shot Performance}\label{sec:zero_shot}

Table~\ref{tab:zero_shot} reports zero-shot performance of all YOLO11 variants on 
the OpenStat Madagascar test set using COCO-pretrained weights without fine-tuning. 
We evaluate the COCO-pretrained weights specifically, rather than an EO-specific 
foundation model, because all our fine-tuned models are initialized from these same 
weights (Section~\ref{sec:methods}); this zero-shot evaluation therefore isolates 
the contribution of our pretraining prior to the fine-tuned performance reported in 
Section~5, rather than benchmarking the best available zero-shot detector for this 
task. All values are near-zero (mAP$_{50} \leq 0.004$ across all models and 
resolutions), confirming the complete absence of exploitable domain transfer from 
COCO pretraining to rooftop PV detection. Larger models do not merely fail to 
benefit from their additional capacity: zero-shot mAP$_{50}$ and recall decrease 
near-monotonically from YOLO11N to YOLO11L (mAP$_{50}$: 0.0022 to 0.0010; recall: 
0.0422 to 0.0096), with only a marginal uptick for YOLO11X. This is consistent with 
the overparameterization penalty documented in Section~5.3: excess capacity appears 
detrimental even before any task-specific fine-tuning occurs.

\begin{table}[h]
  \centering
  \caption{{\bf Zero-shot evaluation of YOLO11 variants on the OpenStat Madagascar
  test set, averaged across available resolutions.} Models use COCO-pretrained
  weights without fine-tuning. F$_1$ is computed from precision and recall.
  All values are near-zero ($\leq 0.004$), confirming the absence of
  domain transfer from COCO pretraining to satellite PV detection}
  \label{tab:zero_shot}
  \setlength{\tabcolsep}{8pt}
  \begin{tabular}{lrrrrr}
    \toprule
    \textbf{Model} & \textbf{mAP$_{50}$} & \textbf{mAP$_{50\text{-}95}$} & \textbf{Precision} & \textbf{Recall} & \textbf{F$_1$} \\
    \midrule
    YOLO11N & 0.0022 & 0.0011 & 0.0041 & 0.0422 & 0.0075 \\
    YOLO11S & 0.0014 & 0.0007 & 0.0027 & 0.0287 & 0.0049 \\
    YOLO11M & 0.0012 & 0.0005 & 0.0024 & 0.0179 & 0.0042 \\
    YOLO11L & 0.0010 & 0.0004 & 0.0018 & 0.0096 & 0.0030 \\
    YOLO11X & 0.0013 & 0.0006 & 0.0024 & 0.0116 & 0.0040 \\
    \bottomrule
  \end{tabular}
\end{table}

\subsection{Resolution and Model Size Interaction}\label{sec:interaction-resolution}
Table~\ref{tab:resolution_model_interaction} reports mean mAP$_{50}$ and efficiency 
per (model, resolution) cell, averaged across dataset fractions. Two distinct patterns 
emerge. In absolute mAP$_{50}$, resolution benefits all models uniformly: YOLO11N 
improves from 0.409 at 416\,px to 0.444 at 1280\,px, while the gap between YOLO11N 
and YOLO11S narrows substantially at 640\,px ($\Delta = 0.008$), suggesting that 
higher resolution partially compensates for the capacity deficit of smaller models 
in raw detection performance. In efficiency, however, the ratio between YOLO11N and 
YOLO11S remains stable at ${\approx}3.5\times$ across all three resolution levels 
(0.80 vs.\ 0.23 at 416\,px; 0.85 vs.\ 0.24 at 640\,px; 0.87 vs.\ 0.25 at 1280\,px): 
resolution scales efficiency proportionally without altering the relative ranking. 
Resolution and model size thus operate on orthogonal axes --- resolution is a universal 
performance lever, but does not compensate for the overparameterization penalty when 
deployment cost is accounted for.
\begin{table}[ht]
\centering
\caption{%
  \textbf{Resolution--model size interaction.}
  Mean mAP\textsubscript{50} (upper block) and mean efficiency
  (mAP\textsubscript{50}\,/\,10\,MB, lower block) per (model, resolution)
  cell, averaged across dataset fractions.
  Bold denotes the best value per column.
  Grey cells indicate configurations excluded due to GPU memory constraints (OOM).
  The efficiency gap between YOLO11N and YOLO11S
  remains stable across resolutions (ratio $\approx 3.5\times$ at all
  three levels), indicating that higher resolution does not compensate
  for the overparameterisation penalty of larger models in terms of
  efficiency. In absolute mAP\textsubscript{50}, the gap narrows
  as the resolution increases}
\label{tab:resolution_model_interaction}
\setlength{\tabcolsep}{7pt}
\begin{tabular}{clrrrrr}
\toprule
& \textbf{Resolution} & \textbf{YOLO11N} & \textbf{YOLO11S} & \textbf{YOLO11M} & \textbf{YOLO11L} & \textbf{YOLO11X} \\
\midrule
\multirow{3}{*}{\rotatebox[origin=c]{90}{\small\textbf{mAP\textsubscript{50}}}} & 416\,px & 0.409 & 0.421 & 0.419 & 0.399 & 0.413 \\
 & 640\,px & 0.433 & 0.441 & 0.436 & \textbf{0.413} & \textbf{0.429} \\
 & 1280\,px & \textbf{0.444} & \textbf{0.448} & \textbf{0.436} & 0.410 & 0.390 \\
\midrule
\multirow{3}{*}{\rotatebox[origin=c]{90}{\small\textbf{Efficiency}}} & 416\,px & 0.80 & 0.23 & 0.11 & 0.08 & 0.04 \\
 & 640\,px & 0.85 & 0.24 & 0.11 & \textbf{0.09} & \textbf{0.04} \\
 & 1280\,px & \textbf{0.87} & \textbf{0.25} & \textbf{0.11} & 0.08 & 0.04 \\
\bottomrule
\end{tabular}
\end{table}
\end{appendix}

\end{Backmatter}

\end{document}